\documentclass{article}
\pdfoutput=1
 \usepackage[preprint]{neurips_2025}
 \usepackage{natbib}

\usepackage[utf8]{inputenc} % allow utf-8 input
\usepackage[T1]{fontenc}    % use 8-bit T1 fonts
\usepackage{hyperref}       % hyperlinks
\usepackage{url}            % simple URL typesetting
\usepackage{booktabs}       % professional-quality tables
\usepackage{amsfonts}       % blackboard math symbols
\usepackage{nicefrac}       % compact symbols for 1/2, etc.
\usepackage{microtype}      % microtypography
\usepackage{xcolor}         % colors
\usepackage{enumitem}

\title{Unlearning as Ablation: Toward a Falsifiable Benchmark for Generative Scientific Discovery}

\author{
  Robert Yang \\
  S6 Research \\
  San Jose, CA 95129 \\
  \texttt{robert@s6research.org} \\
}

\begin{document}

\maketitle

\begin{center}
\textit{Accepted at the NeurIPS 2025 AI4Science Workshop.}
\end{center}

\begin{abstract}
Bold claims about AI’s role in science---from ``AGI will cure all diseases'' to promises of radically accelerated discovery---raise a central epistemic question: do large language models (LLMs) truly \emph{generate} new knowledge, or do they merely remix memorized fragments? We propose \textbf{unlearning-as-ablation} as a falsifiable probe of constructive scientific discovery. The idea is to systematically removes a target result together with its \emph{forget-closure} (supporting lemmas, paraphrases, and multi-hop entailments) and then evaluate whether the model can re-derive the result from only permitted axioms and tools. Success would indicate generative capability beyond recall; failure would expose current limits. Unlike prevailing motivations for unlearning---privacy, copyright, or safety---our framing repositions it as an \emph{epistemic probe} for AI-for-Science. We outline a minimal pilot in mathematics and algorithms to illustrate feasibility, and sketch how the same approach could later be extended to domains such as physics or chemistry. This is a position paper: our contribution is conceptual and methodological, not empirical. We aim to stimulate discussion on how principled ablation tests could help distinguish models that reconstruct knowledge from those that merely retrieve it, and how such probes might guide the next generation of AI-for-Science benchmarks.
\end{abstract}

\section{Introduction}

Recent breakthroughs in foundation models have fueled bold claims---from predictions that ``AGI will cure all diseases'' to assertions that scientific progress will soon accelerate far beyond historical rates. These visions reflect real excitement, but they obscure a fundamental epistemic question: \textbf{do large language models (LLMs) genuinely generate new knowledge, or do they merely remix what was already present in their training data?}

This distinction matters deeply for AI-for-Science. Without a falsifiable test of constructive knowledge generation, claims of ``discovery'' remain philosophically ambiguous and scientifically ungrounded. If AI systems are to be trusted as collaborators in science, we must know whether they can \emph{derive} new results from principles, rather than retrieve or interpolate memorized fragments.

We propose a new perspective: \textbf{unlearning-as-ablation}. The idea is straightforward. Select a scientific result $T$ (e.g., a theorem or algorithm), identify its entire \emph{forget-closure} $\mathcal{F}(T)$---all lemmas, paraphrases, aliases, and multi-hop entailments that lead to $T$---and perform strong unlearning over $\mathcal{F}(T)$. Afterward, provide the model only with permitted axioms and tools, and test whether it can re-derive $T$ in a verifiable form. Success constitutes positive evidence of constructive generation, whereas failure or leakage exposes the boundaries of current capabilities.

This framing departs from prevailing motivations for unlearning. Surveys emphasize privacy, copyright, and safety as primary rationales \citep{5-unlearning-survey, 6-unlearning-survey-2}, with evaluation focused on removal fidelity rather than generative ability. Recent work highlights the difficulty of faithfully removing multi-hop or entangled knowledge \citep{7-multihop, 8-multihop, 9-shah2025the}, while other studies show that forgotten content can often be ``relearned'' through small finetunes or prompting \citep{9-shah2025the}. In the safety and compliance setting, these phenomena are treated as risks. In our setting, they define the frontier: as unlearning methods improve in addressing leakage and robustness, the resulting ablations become more faithful, and the corresponding rediscovery benchmarks more stringent. In this way, progress in unlearning directly strengthens our ability to test whether models are capable of constructive scientific generation.

By reframing unlearning as an \emph{experimental probe}, we aim to bridge AI-for-Science and safety communities. The result is a concrete, falsifiable methodology for testing the limits of LLMs: whether they are capable of genuine discovery, or whether their advances remain bounded by retrieval and interpolation. As a position paper, our contribution is primarily conceptual: we propose a methodological framework and outline pilot domains, leaving systematic empirical validation to future work.

\section{Background: Unlearning Today}

The study of unlearning in machine learning and large language models (LLMs) has grown rapidly in recent years, motivated largely by \emph{external constraints} such as law, safety, or ethics rather than by epistemic goals. We briefly review the dominant rationales, common methodologies, and key evaluation challenges.

\subsection{Motivations for Unlearning}
Three primary motivations recur across surveys and frameworks:

\textbf{(1) Privacy and compliance.} Regulations such as the General Data Protection Regulation (GDPR) enshrine a ``right to be forgotten,'' requiring that models support the removal of sensitive or personally identifiable data. Surveys on digital forgetting in LLMs emphasize compliance with privacy law as a central driver of research in this area \citep{6-unlearning-survey-2}.

\textbf{(2) Copyright and intellectual property.} LLMs trained on large web scrapes may inadvertently memorize copyrighted text, code, or images. Several works argue that machine unlearning is necessary to respect intellectual property claims and to support takedown requests from rights-holders \citep{1-karamolegkou-etal-2023-copyright, 2-copyright-unlearning, 6-unlearning-survey-2, 10-unlearning-survey, 11_NEURIPS2024_be52acf6}.

\textbf{(3) Safety and dual-use knowledge.} A third line of work focuses on removing \emph{hazardous} content: for example, step-by-step instructions for synthesizing explosives or pathogens. Recent benchmarks such as WMDP \citep{12-pmlr-v235-li24bc} evaluate whether unlearning can reduce dual-use risks while maintaining general utility. 

\subsection{Methodological Approaches}
Most unlearning methods adapt techniques from model editing or fine-tuning. Examples include:
\begin{itemize}
    \item \textbf{Gradient-ascent or anti-training:} adjusting model parameters to maximize loss on target examples, thereby forgetting them.
    \item \textbf{Representation-level interventions:} e.g., Amnesic Probing \citep{13-amnesic-probing} removes specific linguistic features from hidden states.
    \item \textbf{Retrieval suppression:} steering methods that block particular outputs without removing underlying representations.
\end{itemize}
While diverse, these approaches generally aim at \emph{removal fidelity}: ensuring that specific facts or behaviors no longer appear in model outputs.

\subsection{Evaluation Challenges}
Evaluation is a persistent bottleneck. Several recent studies emphasize that:
\begin{itemize}
    \item \textbf{Entangled knowledge is difficult to erase.} Multi-hop unlearning benchmarks show that even if intermediate nodes are removed, models can often reconstruct targets via alternative reasoning chains \citep{7-multihop, 8-multihop, 9-shah2025the}.
    \item \textbf{Suppression vs. removal.} SoK papers stress the importance of distinguishing true parameter-level removal from surface-level suppression, where models appear to forget but can be prompted to recall \citep{10-unlearning-survey}.
    \item \textbf{Relearning and robustness.} Empirical work demonstrates that forgotten content can often be ``jogged'' back into use with minimal finetuning or prompting \citep{4-lee2025distillationrobustifiesunlearning}.
\end{itemize}

\subsection{Gap for AI-for-Science}
Notably, none of the above rationales frame unlearning as a tool for \emph{scientific epistemology}. Unlearning has been motivated by compliance and safety, not by the question of whether a model can \emph{reconstruct} forgotten knowledge from first principles. This gap opens an opportunity: by treating unlearning as \emph{ablation}, we can design falsifiable experiments to probe whether LLMs possess constructive generative capabilities, a perspective particularly urgent for AI-for-Science. Moreover, the progress of unlearning research directly determines the strength of such benchmarks: the more thorough and faithful the unlearning, the harder the rediscovery task becomes, and the more reliable the test of whether models can generate knowledge rather than recall it.

\section{Proposal: Unlearning-as-Ablation}

We propose to repurpose unlearning from its conventional role in privacy or safety into an \emph{experimental ablation method} for probing constructive knowledge generation. The central idea is to remove not only a target result $T$, but also all of the \emph{supporting knowledge that directly enables it}, and then ask the model to re-derive $T$ from only axioms and tools that remain accessible. If the model succeeds under these conditions, we gain falsifiable evidence that it is not merely retrieving memorized fragments but genuinely generating knowledge.

\subsection{Defining the Forget-Closure}
The first step is to formally define the \textbf{forget-closure} $\mathcal{F}(T)$ of a target $T$. This closure includes:
\begin{itemize}
    \item All direct statements of $T$ (canonical forms, proofs, code).
    \item Paraphrases and rephrasings that preserve semantic equivalence.
    \item Intermediate lemmas or building blocks that entail $T$.
    \item Multi-hop reasoning chains where $T$ can be reconstructed indirectly \citep{7-multihop, 8-multihop, 9-shah2025the}.
    \item Same-answer sets where multiple formulations yield equivalent outputs.
\end{itemize}
By removing the entire $\mathcal{F}(T)$, we close off not only surface forms but also indirect reasoning paths that would otherwise allow reconstruction through entanglement.

\subsection{Performing Strong Unlearning}
The second step is to apply \textbf{removal-oriented unlearning} across $\mathcal{F}(T)$. Unlike suppression methods that steer generation away from target outputs, removal aims to eliminate relevant information from the parameterization itself. Candidate techniques include gradient-ascent unlearning, targeted fine-tuning, or optimization-based methods evaluated in recent surveys \citep{10-unlearning-survey}. To confirm removal, we propose adopting multi-faceted audits:
\begin{itemize}
    \item Leakage checks on paraphrase, multi-hop, and same-answer sets.
    \item Counterfactual activation probes (inspired by Amnesic Probing) to test whether $T$-related features still reside in hidden states \citep{13-amnesic-probing}.
    \item Robustness tests against ``jogging'' attacks, where small finetunes or prompting can restore forgotten knowledge \citep{4-lee2025distillationrobustifiesunlearning}.
\end{itemize}
These checks ensure that the unlearning process produces a genuine epistemic blank slate with respect to $\mathcal{F}(T)$.

\subsection{Re-Derivation as a Falsifiable Test}
Finally, we design a \textbf{re-derivation trial}. After unlearning, the model is provided with:
\begin{enumerate}
    \item A set of axioms, primitives, or base tools that are \emph{not} part of $\mathcal{F}(T)$.
    \item A prompt or environment that permits constructive reasoning (e.g., a proof assistant or a test-driven code synthesis framework).
\end{enumerate}
The task is to derive $T$ in a form that can be verified by an external oracle: for example, a formal proof accepted by Lean or Isabelle, or a program passing a hidden test suite. Importantly, success is only counted if $T$ is re-derived \emph{without leakage from $\mathcal{F}(T)$}. 

This yields a falsifiable criterion: if the model can re-derive $T$ despite rigorous unlearning of all prerequisite paths, we have positive evidence for constructive generation. If it cannot, or if leakage audits reveal dependence on residual memory, then the claim of ``scientific discovery'' remains unsubstantiated.

\subsection{Why This Matters}
This approach connects progress in unlearning directly to progress in measuring scientific discovery. In the safety and compliance literature, challenges such as entanglement, multi-hop reasoning, and relearning are treated as failure modes because they undermine removal fidelity \citep{2-copyright-unlearning,7-multihop,8-multihop,9-shah2025the}. In our framing, they set the difficulty of the benchmark: the more effectively unlearning methods address these challenges, the more thoroughly the target knowledge is ablated, and the more demanding the rediscovery task becomes. Thus, advances in unlearning translate into sharper tests of whether LLMs truly possess constructive generative capability. Rather than turning flaws into benefits, we highlight that solving these long-standing problems in unlearning is what enables rigorous epistemic evaluation in AI-for-Science.

\section{Minimal Pilot Study}

While the long-term vision is to apply unlearning-as-ablation to scientific hypotheses in physics, chemistry, or biology, we propose beginning with domains where \textbf{verification is automatic and unambiguous}. This allows us to isolate the epistemic question---can a model \emph{re-derive} knowledge once its closure has been forgotten?---without relying on subjective human judgment.

\subsection{Mathematics: Formal Proofs}
Mathematics provides an ideal testbed because results can be verified by proof assistants such as \emph{Lean} or \emph{Isabelle}. A minimal pilot could proceed as follows:
\begin{enumerate}
    \item Select a mid-tier theorem (e.g., in number theory or combinatorics) that has a clear dependency structure.
    \item Construct its forget-closure $\mathcal{F}(T)$, including canonical statements, paraphrased variants, and prerequisite lemmas.
    \item Apply strong unlearning over $\mathcal{F}(T)$.
    \item Task the model with re-proving $T$ using only base axioms and allowed rules of inference.
\end{enumerate}
Success is defined as producing a proof accepted by the proof assistant. Failure or leakage (e.g., shortcut recall of a forgotten lemma) falsifies the claim of rediscovery.

\subsection{Algorithms: Verified Implementations}
Algorithms provide another tractable domain, where correctness can be checked against hidden test suites. For example:
\begin{enumerate}
    \item Forget the Knuth--Morris--Pratt (KMP) string matching algorithm, along with all prerequisite explanations, code templates, and paraphrases.
    \item After unlearning, ask the model to derive an efficient string-matching procedure from first principles (e.g., reasoning about prefix functions).
    \item Validate correctness using adversarial test cases and runtime complexity checks.
\end{enumerate}
As in mathematics, the evaluation is binary: either the model reconstructs a working implementation, or it does not.

\subsection{Evaluation Metrics}
To assess the outcome of such pilots, we propose three classes of metrics:
\begin{itemize}
    \item \textbf{Success rate.} Fraction of trials where the model re-derives $T$ in a verifiable form (proof acceptance, program passes test suite).
    \item \textbf{Leakage audits.} Performance on paraphrase, multi-hop, and same-answer sets drawn from $\mathcal{F}(T)$, ensuring the model is not recalling forgotten material \citep{7-multihop, 8-multihop, 9-shah2025the}.
    \item \textbf{Utility retention.} Accuracy on unrelated benchmarks (e.g., a subset of MMLU) to confirm that unlearning did not degrade general capability \citep{10-unlearning-survey, 11_NEURIPS2024_be52acf6}.
\end{itemize}

\subsection{Why a Minimal Pilot is Valuable}
Even small-scale pilots can decisively answer whether LLMs exhibit generative capability under ablation. If a model successfully re-derives a theorem or algorithm after strong unlearning of its closure, we obtain falsifiable evidence that it constructs knowledge rather than merely retrieving it. Conversely, if models fail under such controlled conditions, this highlights a concrete epistemic limit of current systems. Either outcome offers high-value insight for AI-for-Science, where claims of accelerated discovery remain both enticing and contested.

\section{Implications for AI-for-Science}

The proposed unlearning-as-ablation framework has direct consequences for how we understand the promise and limits of AI-for-Science.

\subsection{Epistemic Clarity in Scientific Discovery}
The central value of this approach is that it provides a \emph{falsifiable test} of discovery. Today, when an LLM proposes a hypothesis, proves a theorem, or writes an algorithm, it remains unclear whether this is a product of genuine reasoning or of subtle retrieval from training data. By first \emph{removing} all accessible pathways to a result and then testing for \emph{re-derivation}, we create a clean epistemic separation: success implies constructive generation, while failure implies dependence on stored fragments. This reframing allows the AI-for-Science community to move beyond speculation about ``discovery'' and instead ground claims in falsifiable evidence.

\subsection{Turning Failure Modes into Probes}
Unlearning research has traditionally cast entanglement, multi-hop reasoning, and relearning as obstacles \citep{7-multihop,8-multihop,9-shah2025the}. In our setting, these challenges become useful stress tests. If a model cannot succeed once closure paths are blocked, it indicates that the relevant knowledge was never truly generative. If it can succeed, it demonstrates robustness and constructive capacity. Either way, phenomena previously treated as evaluation headaches become diagnostic instruments for probing the depth of model reasoning.

\subsection{Broader AI-for-Science Roadmap}
Although we highlight mathematics and algorithms as tractable pilot domains, the methodology generalizes. In physics, one could remove an established equation and test whether the model can re-derive it from fundamental laws. In chemistry, one could unlearn a well-known synthesis route and test whether the model can rediscover it from reaction rules. In biology, one could unlearn a canonical protein interaction and test for re-derivation from structural principles. These extensions would demand careful closure construction and domain-specific verification, but they illustrate how the same ablation logic scales to real scientific practice.

\subsection{Redefining the Boundary of AI Progress}
Finally, this framework speaks directly to the theme of this workshop: the reach and limits of AI in scientific discovery. If unlearning-as-ablation pilots reveal that models can re-derive knowledge under strong ablation, this strengthens the case that AI can generate truly novel insights. If they reveal consistent failures, it delineates a boundary condition: LLMs may accelerate retrieval, interpolation, and synthesis, but fall short of independent knowledge generation. In both outcomes, the methodology provides a principled way to map the contours of what AI can and cannot do for science. 

\subsection{Toward the Next Major Benchmark}
A final implication is that unlearning-as-ablation offers a clear path toward the next generation of benchmarks for AI progress. Just as ImageNet catalyzed advances in computer vision by providing a well-defined task on which algorithms could be compared \citep{15-deng2009imagenet}, a benchmark grounded in constructive re-derivation after unlearning could serve as a lodestar for AI-for-Science. Existing evaluations of knowledge regurgitation and short-form reasoning are increasingly saturated---as highlighted by works such as Humanity's Last Exam \citep{14-phan2025humanitysexam}---suggesting that the next frontier must measure whether models can move beyond retrieval and interpolation to genuine discovery. We believe that such an ``unlearning-as-ablation'' benchmark could become a distinguishing test of model strength, separating systems that can merely recall from those that can constructively generate new scientific knowledge.

Importantly, the strength of such a benchmark is coupled to the progress of unlearning research itself. As unlearning methods become more faithful and thorough, the corresponding benchmarks become more stringent: rediscovery requires deeper reasoning, and successful re-derivation provides stronger evidence of constructive capability. In this way, advances in unlearning directly drive advances in our ability to measure---and eventually to achieve---genuine AI scientific discovery.

\section{Conclusion}

We have proposed \emph{unlearning-as-ablation} as a new lens on large language models, reframing unlearning from a tool of compliance and safety into a falsifiable probe of scientific discovery. By systematically removing a target result and its forget-closure, and then testing whether the model can re-derive the result from permitted axioms and tools, we obtain an experimental method to separate retrieval from constructive generation. This approach directly addresses one of the most pressing open questions in AI-for-Science: can AI systems truly generate new knowledge? Even minimal pilots in mathematics or algorithms provide decisive evidence either way, while extensions to physics, chemistry, and biology can delineate the boundaries of future AI scientific progress. Whether the outcome is success or failure, unlearning-as-ablation offers the community a principled framework to move beyond speculation and anchor claims of discovery in falsifiable tests.

\section{Transparency on AI usage}
Although not required for NeurIPS submission, for full transparency of the preprint, we include this disclosure here. AI was used extensively throughout the paper for editing (suggesting terminology use and phrasing, plus paraphrasing of the author's writing to increase quality) fully adhering to NeurIPS 2025 guidelines. The authors are responsible for the entire content of the paper, including all text, figures, and references.

\bibliographystyle{plainnat}
\bibliography{refs}

\appendix

\section{Benchmark Specification}

We propose Ablation-to-Discovery (A2D), a benchmark that frames unlearning as ablation to test whether large language models can reconstruct systematically removed knowledge from first principles. By defining rediscovery tasks with verifiable outcomes, A2D probes constructive generative ability—can models re-derive what was excised? This offers a falsifiable substrate for evaluating knowledge generation beyond memorization. Just as ImageNet galvanized computer vision, we envision A2D as the “ImageNet of knowledge generation”—a shared testbed for measuring and accelerating AI-for-Science progress.

\subsection{Dataset Rationale — Why an Ablation-Coupled Benchmark?}

LLMs are saturated on recall-heavy tasks but under-tested on constructive generation. A2D provides a controlled falsifiable test: remove structured knowledge $T$, then evaluate if models can rebuild it without rote recall.

Unlearning is typically framed as a risk mitigation strategy (safety, privacy) \citep{pmlr-v235-huang24u, 6-unlearning-survey-2, 10-unlearning-survey}. Here, we reframe it as a methodological opportunity: each advance in unlearning methods strengthens ablations, raising the difficulty of reconstructive discovery. Thus, progress in unlearning directly drives progress in A2D benchmarks.

These are our key rationales:

\begin{itemize}[itemsep=0pt, topsep=0pt, parsep=0pt, partopsep=0pt]

\item Scientific falsifiability: A2D enables yes/no tests of generative capability.

\item Reproducibility: Each task ships as a containerized config.

\item Benchmark trajectory: Initial domains (math, algorithms), then expansion into physics, chemistry, biology, and other basic sciences.

\item Community role: A2D can serve as a battleground for AI Scientist frameworks, providing the first quantitative ground for comparing systems like Google’s biotech discovery AI \citep{gottweis2025ai_co_scientist} or Sakana’s automated CS paper generation \citep{lu2024ai_scientist, yamada2025ai_scientist_v2}.

\end{itemize}

\subsection{AI Task Definition}

Traditional benchmarks equate “dataset” with a static collection of labeled examples—ImageNet, GLUE, and many others embody this paradigm \citep{deng2009imagenet, wang2018glue}. Our proposal expands this definition. In the unlearning-as-ablation setting, the benchmark is not merely the data but the procedure by which knowledge is systematically removed. In some cases, it also encompasses a standardized reference model that undergoes ablation. In this view, a dataset is no longer just an archive of examples, but a dynamic specification of data, process, and model.

This redefinition is essential. By treating ablation as part of the benchmark, we can directly test whether models or systems can reconstruct algorithmic rules, scientific knowledge, or cross-domain mappings that have been deliberately removed. Without embedding the ablation protocol (and in some cases, the model artifact) into the benchmark, such generative reconstruction cannot be meaningfully evaluated. Thus, our task definition is broader than “predict labels for examples.” It is: given a systematically ablated knowledge space, recover the missing structure with scientific fidelity.

\subsection{Tracks and Modes}
Our proposal separates evaluation along two orthogonal dimensions: tracks and modes.

\paragraph{(1) Tracks (what is reconstructed):}

\begin{itemize}[itemsep=0pt, topsep=0pt, parsep=0pt, partopsep=0pt]

\item Algorithmic Re-derivation – rediscovering hidden formal rules or procedures.

\item Scientific Knowledge Reconstruction – restoring ablated domain-specific knowledge (e.g., molecular pathways, physics laws) by reasoning from foundational laws and experimental constraints?

\item Cross-Domain Generalization – leveraging one domain to recover knowledge in another. For example, can a model, after relevant results are removed, re-derive a computational biology method by combining algorithmic and biochemical principles?

\end{itemize}

\paragraph{(2) Evaluation Modes (how the test is administered):}

\begin{itemize}[itemsep=0pt, topsep=0pt, parsep=0pt, partopsep=0pt]

\item BYOM Capability Test – the benchmark specifies only the ablation protocol, with models tested directly (no discovery frameworks), to isolate capability.

\item System Capability Test – the benchmark includes both the ablation protocol and a standardized reference model artifact. Here, the comparison is among discovery frameworks, evaluating how orchestration, augmentation, or agentic processes recover knowledge.

\end{itemize}

By separating Task Tracks from Evaluation Modes, the benchmark distinguishes between what kind of knowledge generation is being probed and whether the rediscovery is attributable to the model alone or to a composite system.

\subsection{Acceleration Potential — Unlocking Constructive AI-for-Science}

\paragraph{Catalyzing a new benchmark frontier:} As ImageNet did for vision \citep{deng2009imagenet}, A2D offers a single ground truth task for constructive scientific discovery.

\paragraph{Driving a virtuous cycle:} Stronger unlearning leads to stronger ablations, meaning harder benchmarks, which drives sharper evaluation of generative capacity.

\paragraph{Serving as battleground for AI Scientist frameworks:} Recent “AI Scientist” efforts (e.g. Google’s biotech discovery \citep{gottweis2025ai_co_scientist}, Sakana AI’s paper generation \citep{lu2024ai_scientist, yamada2025ai_scientist_v2}) demonstrate ambition but lack common evaluation. A2D provides the first quantifiable arena for comparing them.

\paragraph{Impact:} 1) Establishes a rigorous test for constructive generative ability; 2) Accelerates AI-for-Science by standardizing falsifiable evaluation; 3) Offers rapid adoption via lightweight, containerized tasks.

\section{Extended Benchmark Specification}

\subsection{Extended Draft on Dataset Rationale — Why an Ablation-Coupled Benchmark?}

\paragraph{The bottleneck.} AI has advanced in waves catalyzed by benchmarks: ImageNet for vision \citep{deng2009imagenet}, Common Crawl for pretraining \citep{commoncrawl}, and MMLU or Humanity’s Last Exam for reasoning \citep{hlx2024}. Today, models already saturate benchmarks based on knowledge regurgitation and short-form reasoning. What remains unmeasured is the ability to reconstruct forgotten knowledge from first principles. Without such a test, claims that AI systems make genuine scientific discoveries cannot be falsified. Thus, the bottleneck is not simply data volume, but the absence of a dataset that (i) ensures controlled forgetting and (ii) provides automatic verification of rediscovery.

\paragraph{What the dataset consists of.}
Our dataset proposal, \textbf{Ablation-to-Discovery (A2D)}, is defined by triplets of:

\begin{itemize}[itemsep=0pt, topsep=0pt, parsep=0pt, partopsep=0pt]

\item Target specification ($T$): a theorem, algorithm, or identity stated in a machine-checkable form.

\item Forget-closure ($\mathcal{F}(T)$): a structured collection of all paraphrases, prerequisite lemmas, aliases, and multi-hop derivations that entangle with $T$. Each closure comes with paraphrase/multi-hop/same-answer probes to audit leakage.

\item Ablation recipe ($\mathcal{A}(T)$): a reproducible pipeline that, given a base checkpoint, produces an ablated checkpoint in which $\mathcal{F}(T)$ is unlearned to a specified fidelity.

\end{itemize}

Each instance also includes a verification oracle ($\mathcal{V}(T)$) (proof assistant kernel, hidden program test suite, or physics constraint checker) to determine whether the model’s output constitutes a valid re-derivation.

\paragraph{Scale and scope.}

\begin{itemize}[itemsep=0pt, topsep=0pt, parsep=0pt, partopsep=0pt]

\item Initial release: 50–100 pilot instances across mathematics and algorithms, where verification is automatic and the dependency graphs are tractable.

\item Growth path: community contributions of new $T$ and $\mathcal{F}(T)$ pairs in physics, chemistry, and biology. These can scale into hundreds or thousands of benchmark items over time, analogous to the growth of ImageNet categories \citep{deng2009imagenet}.

\item Resolution and metadata: each item is richly annotated with dependency graphs, paraphrase sets, ablation configs, and verification schemas—making it reusable for both unlearning and discovery research.

\end{itemize}

\paragraph{Why existing datasets are insufficient.}

\begin{itemize}[itemsep=0pt, topsep=0pt, parsep=0pt, partopsep=0pt]

\item Knowledge editing datasets (e.g., MEMIT \citep{meng2022memit}, ROME \citep{zhou2023rome}) test whether models can adjust facts, but do not couple deletion with generative rediscovery.

\item Safety benchmarks (e.g., WMDP \citep{li2024wmdp}) test suppression of hazardous knowledge, but not constructive derivation.

\item Reasoning benchmarks (MMLU \citep{hendrycks2020mmlu}, GSM8K \citep{cobbe2021gsm8k}) test regurgitation or short reasoning chains, but not reconstruction after ablation.

\end{itemize}

\paragraph{Why ablation must be part of the dataset.} If only the target questions $T$ were included, results would be confounded by uncontrolled leakage from pretraining corpora. By including \emph{the ablation recipes themselves} as part of the dataset, every researcher can reproduce equivalent epistemic conditions. In this way, the dataset defines not just the task, but the controlled \emph{epistemic starting point} for fair comparison across models and systems.

\subsection {Extended Draft on AI Task Definition}

% $\mathcal{F}(T)$
\paragraph{Core scientific question.} 
Can an AI system constructively re-derive a target scientific result $T$ (e.g., theorem, algorithm, physical identity) after the model has been systematically unlearned of $T$
and its forget-closure $\mathcal{F}(T)$
(all lemmas, paraphrases, templates, and multi-hop entailments that enable $T$)? This is a generation task with external verification (formal proof acceptance or program/test-suite pass), explicitly designed to distinguish retrieval/interpolation from genuine derivation.

\paragraph{Benchmark instances (“tasks”).}
Each instance packages four components:

\begin{itemize}[itemsep=0pt, topsep=0pt, parsep=0pt, partopsep=0pt]

\item Target spec $T$: a formally stated goal (e.g., Lean theorem, algorithmic spec, physics identity).

\item Closure spec $\mathcal{F}(T)$: machine-readable lists/patterns for direct statements, paraphrases, prerequisite lemmas, multi-hop chains, and same-answer sets.

\item Ablation recipe $\mathcal{A}(T)$: a reproducible unlearning pipeline (config + seed) that takes a base model checkpoint and outputs an ablated checkpoint in which $\mathcal{F}(T)$ is removed to a specified fidelity threshold.

\item Verification oracle $\mathcal{V}(T)$: an automatic checker (e.g., Lean/Isabelle kernel; hidden program tests; executable physics constraints) that returns accept/reject and auxiliary traces.

\end{itemize}

\paragraph{Task input.} An ablated model (produced by running $\mathcal{A}(T)$ on a supported base model), the allowed axioms/tools (e.g., proof-assistant primitives, standard libraries specified by the task), and the target spec $T$ (no examples or templates from $\mathcal{F}(T)$).

\paragraph{Task output.}
A candidate derivation of $T$: a formal proof that $\mathcal{V}(T)$ accepts (math/logic tracks), or an artifact (program/spec) that $\mathcal{V}(T)$ validates against hidden tests (algorithms/physics/chemistry tracks).

\paragraph{Why the ablation is part of the benchmark.}
Model comparisons are only fair if knowledge leakage is controlled. Treating the ablation pipeline as first-class data ensures every submission is evaluated under equivalent epistemic conditions. (We will also provide reference ablated checkpoints for popular base models to enable system-level, apples-to-apples comparisons.)

\paragraph{Modes (two complementary comparison modes).}

\begin{itemize}[itemsep=0pt, topsep=0pt, parsep=0pt, partopsep=0pt]

\item \textbf{Mode A - Model/Agent Mode (Bring-Your-Own Model).} Participants run the provided $\mathcal{A}(T)$ on their model, then attempt re-derivation using only allowed tools. This mode is used to compare model performance.

\item \textbf{Mode B - System/Framework Mode (Standardized Model).} Participants use provided, fixed ablated checkpoints (e.g., "A2D-Llama-X-Ablated-v1") to compare science-discovery frameworks (planners, tool-use agents, proof searchers) independent of pretraining.

\end{itemize}

\paragraph{Primary metric.} Pass@k on $\mathcal{V}(T)$ (e.g., proof acceptance or full test pass) with strict time/compute budgets per instance.

\paragraph{Secondary diagnostics.}

\begin{itemize}[itemsep=0pt, topsep=0pt, parsep=0pt, partopsep=0pt]
\item Leakage audits (paraphrase/multi-hop/same-answer probes defined in $\mathcal{F}(T)$);

\item Robustness (success stability under small prompt/seed changes);

\item Efficiency (wall-clock, tool calls) under fixed budgets.
\end{itemize}

\paragraph{Roadmap}

\begin{itemize}[itemsep=0pt, topsep=0pt, parsep=0pt, partopsep=0pt]

\item \textbf{Initial domains: } Mathematics (Lean/Isabelle-verifiable theorems); Algorithms (spec-driven implementations with hidden adversarial tests).

\item \textbf{Road-map domains (as community contributions mature):} Physics identities/constraints, chemical synthesis steps, and biology mechanisms with simulators or curated oracles.

\end{itemize}

\paragraph{Reproducibility and shareability.} 

\begin{itemize}[itemsep=0pt, topsep=0pt, parsep=0pt, partopsep=0pt]
\item All instances ship as containers with $\mathcal{A}(T)$, $\mathcal{V}(T)$, and JSON schemas for $T$/$\mathcal{F}(T)$;

\item Seeded runs; deterministic configs; checksum’d ablated checkpoints for Mode B;

\item Licensing and redistribution policies aligned with base-model terms.
\end{itemize}

\subsection{Extended Draft on Acceleration Potential — Unlocking Constructive AI-for-Science}

\paragraph{Catalyzing a new benchmark frontier.}
The Ablation-to-Discovery (A2D) dataset would establish the first falsifiable benchmark for constructive scientific generation. Just as ImageNet provided a hill-climbable substrate that fueled deep learning in vision \citep{deng2009imagenet}, A2D would let researchers systematically compare models and architectures on their ability to re-derive knowledge once its closure has been forgotten. By defining both the targets and the ablation process, A2D transforms “scientific discovery” from a vague aspiration into a concrete, measurable capability.

\paragraph{Driving unlearning and discovery in tandem.}
Progress in unlearning directly amplifies the challenge of A2D: the more faithfully $\mathcal{F}(T)$ is removed, the harder the rediscovery task becomes, and the more diagnostic success becomes. This coupling ensures a virtuous cycle: advances in unlearning sharpen the benchmark, which in turn forces advances in reasoning, derivation, and discovery frameworks.

\paragraph{Impact on model development.}

\begin{itemize}[itemsep=0pt, topsep=0pt, parsep=0pt, partopsep=0pt]

\item For foundation model developers, A2D provides a rigorous testbed for epistemic capability: beyond pass rates on factual recall, can a model constructively rebuild forgotten results?

\item For system builders (e.g., AI scientists, tool-augmented agents), A2D offers a standardized arena where strategies for exploration, reasoning, and tool use can be fairly compared—either by running ablation on their own models or by using standardized ablated checkpoints.

\item For evaluation researchers, A2D creates a new class of benchmarks that integrate unlearning fidelity, rediscovery performance, leakage audits, and utility retention.

\end{itemize}

\paragraph{Cross-domain acceleration.}
While the initial release focuses on mathematics and algorithms (where verification is strict and automatic), the same paradigm extends naturally:

\begin{itemize}[itemsep=0pt, topsep=0pt, parsep=0pt, partopsep=0pt]

\item Physics: unlearn an equation, test rediscovery from fundamental laws.

\item Chemistry: unlearn a synthesis pathway, test rediscovery from reaction rules.

\item Biology: unlearn a canonical interaction, test rediscovery from structural constraints.

\end{itemize}

Each new domain added to A2D increases its reach, creating a shared platform where diverse scientific communities can evaluate constructive AI progress under consistent epistemic conditions.

\paragraph{Rapid, widespread impact.}
Because A2D instances are modular and reproducible (target + closure + ablation recipe + oracle), the benchmark can be shared openly and extended collaboratively. New models, architectures, and discovery frameworks can be stress-tested immediately. This positions A2D to become a community-wide standard for measuring the one capability that matters most for AI-for-Science: moving beyond retrieval to genuine discovery.

\subsection{Extended Draft on Data-Creation Pathway}

The core of A2D is not a single static dataset but a reproducible protocol: select a scientific target $T$, apply an ablation procedure $\mathcal{A}(T)$ to a base model $M$, and record the model’s attempt to rediscover $T$. This shifts the notion of “data creation” from raw collection to repeatable transformation.

Concretely, the pathway looks like this:

\begin{itemize}[itemsep=0pt, topsep=0pt, parsep=0pt, partopsep=0pt]

\item Target selection: Curators provide a library of canonical scientific theorems, proofs, or results (e.g., Euler’s formula, Mendel’s laws, Maxwell’s equations).

\item Ablation recipes: For each target $T$, a documented recipe specifies how to apply an unlearning or fine-tuning procedure that removes $T$ from $M$.

\item Rediscovery logs: Researchers run their system on the ablated model, generating traces of attempted rediscovery (reasoning chains, intermediate hypotheses, final answers). These logs constitute the comparable benchmark outputs.

\end{itemize}

Because recipes and protocols are public, the pathway is scalable and decentralized. Researchers can regenerate ablated models locally or use shared checkpoints for convenience. The “dataset” is thus partly static (targets, configs, evaluation scripts) and partly dynamic (rediscovery logs generated under standardized conditions).

Looking forward, we expect this pathway to become even more streamlined. Emerging infrastructures for model editing and unlearning—potentially delivered through “Ablation-as-a-Service” platforms—could automate recipe application, verification, and distribution. In the longer term, agentic pipelines might automatically curate new scientific targets, generate validated ablation configs, and integrate them into the benchmark with minimal human oversight. This vision makes A2D not just a dataset but a self-renewing ecosystem, capable of expanding alongside advances in both science and AI.

\subsection{Extended Draft on Cost and Scalability}

The primary new cost introduced by A2D lies in generating ablated models. Unlike conventional benchmarks, where fixed datasets can be distributed once, A2D requires creating model variants with targeted unlearning. This raises the question of whether the benchmark is too expensive to scale.

In practice, the cost is modest. Producing an ablated model typically requires tens to hundreds of GPU-hours of unlearning, orders of magnitude less than the millions of GPU-hours consumed by foundation model pretraining. Moreover, the benchmark is defined by protocols rather than a static zoo of checkpoints. Ablation recipes $\mathcal{A}(T)$ can be published alongside base-model identifiers, enabling any researcher to reproduce ablated variants locally. This shifts the cost structure from centralized curation to distributed, on-demand regeneration.

For adoption, two models of distribution are possible. At minimum, benchmark curators can release ablation configs and evaluation scripts, minimizing central cost. For convenience, reference ablated checkpoints can also be shared, incurring modest additional compute and storage but lowering the barrier to entry. Either way, the marginal cost of scaling A2D is low, with the community’s effort concentrated not on compute but on validating that ablations are faithful and consistent.

Looking forward, the cost trajectory is favorable: as unlearning techniques become more efficient and standardized, the marginal expense of producing ablations will fall. Emerging toolkits and infrastructure—potentially “Unlearning-as-a-Service”—could make generating ablated models nearly as routine as dataset preprocessing, further lowering barriers and enabling broader community participation.

\section{Author's Final Remarks}

Knowledge is a lottery ticket for technological advancement. In the book "Why Greatness Cannot Be Planned" by \cite{stanley2015why}, the authors conjecture that "[a]lmost no prerequisite to any major invention was invented with that invention in mind". Rather, it is rather unclear how even to assemble the prerequisites to great inventions before they are invented.

To increase the chances of the next major technological breakthrough happening within our lifetimes, we therefore need to increase the volume at which "intellectual novelties" (knowledge) is being generated. Artificial intelligence is one of the ways toward this goal; in a way, we can consider it as a "lottery ticket printer"; it prints out the numbers and we just have to verify whether we have found the jackpot.

Furthermore, interdisciplinary connectivity does not scale linearly with the amount of available knowledge. As the number of distinct ideas, tools, and domains increases, the number of potential cross-domain linkages grows on the order of the square of that number. Because many breakthroughs arise from previously unanticipated combinations of concepts, this combinatorial expansion implies that the probability of encountering a useful synthesis may grow superlinearly—potentially polynomially rather than linearly—with the total volume of knowledge.

The remaining bottleneck is the reliability of the “lottery tickets” being generated. An unreliable generator may propose nonexistent, incoherent, or systematically incomplete possibilities, limiting the effective search space regardless of volume. The framework developed in this paper aims to eventually produce the environment needed to shape the conditions under which knowledge-generation systems produce increasingly faithful, comprehensive, and well-calibrated hypotheses. In doing so, it moves these “lottery-ticket printers” toward scale and reliability.

\end{document}